\documentclass{article}
\pdfoutput=1

\usepackage{arxiv}

\usepackage[utf8]{inputenc} 
\usepackage[T1]{fontenc}    
\usepackage{hyperref}       
\usepackage{url}            
\usepackage{booktabs}       
\usepackage{amsfonts}       
\usepackage{pifont}
\usepackage{nicefrac}       
\usepackage{microtype}      
\usepackage{lipsum}
\usepackage{graphicx}
\usepackage{amsmath}
\usepackage{multirow}
\usepackage{multicol}
\usepackage{subfigure}

\usepackage[ruled,linesnumbered]{algorithm2e}
\graphicspath{ {./images/} }

\title{Adaptive Graph Diffusion Networks}

\author{
 Chuxiong Sun \\
  Big Data and AI Division\\
  China Telecom Research Institute \\
  Beijing, China \\
  \texttt{chuxiongsun@gmail.com} \\
 \And
 Jie Hu \\
  Big Data and AI Division\\
  China Telecom Research Institute \\
  Beijing, China \\
  \texttt{hujie1@chinatelecom.cn} \\
 \And
 Hongming Gu \\
  Big Data and AI Division\\
  China Telecom Research Institute \\
  Beijing, China \\
  \texttt{guhm@chinatelecom.cn} \\
 \And
 Jinpeng Chen \\
  School of Software Engineering\\
  Beijing University of Posts and Telecommunications\\
  Beijing, China \\
  \texttt{jpchen@bupt.edu.cn} \\
 \And
 Mingchuan Yang \\
  Big Data and AI Division\\
  China Telecom Research Institute \\
  Beijing, China \\
  \texttt{yangmch@chinatelecom.cn} \\

}

\begin{document}
\maketitle
\begin{abstract}
Graph Neural Networks (GNNs) have received much attention in the graph deep learning domain. However, recent research empirically and theoretically shows that deep GNNs suffer from over-fitting and over-smoothing problems. The usual solutions either cannot solve extensive runtime of deep GNNs or restrict graph convolution in the same feature space. We propose the Adaptive Graph Diffusion Networks (AGDNs) which perform multi-layer generalized graph diffusion in different feature spaces with moderate complexity and runtime. Standard graph diffusion methods combine large and dense powers of the transition matrix with predefined weighting coefficients. Instead, AGDNs combine smaller multi-hop node representations with learnable and generalized weighting coefficients. We propose two scalable mechanisms of weighting coefficients to capture multi-hop information: Hop-wise Attention (HA) and Hop-wise Convolution (HC). We evaluate AGDNs on diverse, challenging Open Graph Benchmark (OGB) datasets with semi-supervised node classification and link prediction tasks. Until the date of submission (Aug 26, 2022), AGDNs achieve \textbf{top-1} performance on the ogbn-arxiv, ogbn-proteins and ogbl-ddi datasets and \textbf{top-3} performance on the ogbl-citation2 dataset. On the similar Tesla V100 GPU cards, AGDNs outperform Reversible GNNs (RevGNNs) with 13\% complexity and 1\% training runtime of RevGNNs on the ogbn-proteins dataset. AGDNs also achieve comparable performance to SEAL with 36\% training and 0.2\% inference runtime of SEAL on the ogbl-citation2 dataset.
\end{abstract}



\section{Introduction}

The Graph Neural Networks (GNNs), or Message Passing Neural Networks (MPNNs) \cite{gilmer2017neural}, recently proved effective and became mainstream of graph deep learning in many domains, such as citation networks \cite{kipf2016semi, hamilton2017inductive, kipf2016variational}, social networks \cite{hamilton2017inductive, chen2018fastgcn}, biological graphs \cite{fout2017protein}, and traffic networks \cite{yu2021deep, li2021traffic, yin2021multi}. Based on simple feed-forward networks, they perform message passing in each layer to capture neighborhood information. However, unlike the deep models in the Computer Vision (CV) domain, the deep GNNs \cite{li2018deeper, wang2019improving} will encounter the over-smoothing problem, resulting in significant performance degradation. The model cannot distinguish the node representations, which become nearly identical after long-range message passing. Deep GNNs may contain many feature transformations since a GNN layer usually couples a graph convolution operator and a feature transformation. Recent research \cite{liu2020towards} reveals that the well-known overfitting problem caused by redundant transformations can contribute significantly to the performance degradation of deep GNNs. In addition, many transformations also bring high memory footprints and extended runtime.


By emphasizing the shallow information during long-range message passing, several types of residual connections \cite{li2018deeper, li2019deepgcns, li2020deepergcn, chen2020simple,li2021training} can tackle the over-smoothing and over-fitting problems. However, they do not improve extensive high memory footprints or runtime from deep GNNs. Some techniques \cite{li2021training} from the CV domain can effectively reduce memory footprints of deep GNNs, but with a longer runtime.

Residual connections work among pairs of graph convolution and feature transformation. In contrast, the graph diffusion-based methods directly combine sequential graph convolution operators. The shallow information can be preserved with suitable weighting coefficients to alleviate the oversmoothing problem. A graph convolution operator can be described as a matrix multiplication between the weighted adjacency (transition matrix) and node feature matrix. Then a graph diffusion operator replaces the transition matrix with a linear combination of its different powers.
On the other hand, since many feature transformations bring both efficiency and performance problems, limiting the number of transformations is also helpful. Thus, employing the graph diffusion with limited feature transformations is reasonable. 

This strategy exists in some decoupled methods \cite{liu2020towards, klicpera2018predict, rossi2020sign} that decouple graph convolution operators and feature transformations. In detail, all graph convolution operators are placed before or after all feature transformations. Thus, the decoupled GNNs can enlarge the receptive field with shallow feature transformations. The graph diffusion or its variants are incorporated to leverage the multi-hop information. However, there are no feature transformations (non-linear hierarchies) among all graph convolution operators. Graph convolution is restricted in the same feature space. This characteristic may limit their model capacity. For example, decoupled GNNs can achieve considerable performance on small citation datasets but residual GNNs outperform them on the larger ogbn-arxiv dataset. 

Directly replacing the graph convolution operator in each GNN layer with the graph diffusion operator can also enlarge the receptive field with shallow feature transformations. Graph Diffusion Convolution (GDC) directly replaces the transition matrix with an explicit diffusion matrix. Efficient approximation and sparsification techniques can reduce the memory cost of the dense and large diffusion matrix. However, GDC lacks flexibility since its weighting coefficients are predefined and fixed for different nodes and feature channels. Moreover, GDC cannot improve link prediction performance. A more efficient but equivalent way of calculating the graph diffusion is to calculate multi-hop node representations iteratively and combine them. Some methods perform this memory-efficient graph diffusion in each layer without calculating the explicit diffusion matrix. In other words, instead of high-power transition matrices, multi-hop representation matrices are calculated and stored. The unified framework of these methods has not been well studied and separated from other graph diffusion-based methods \cite{klicpera2018predict, liu2020towards, klicpera2019diffusion}. We summarize them as Graph Diffusion Networks (GDNs). However, many existing GDNs still utilize the fixed predefined hop weighting coefficients, which are identical across nodes, channels, and layers.

Other techniques \cite{zhao2019pairnorm, rong2019dropedge, feng2020graph} that do not modify model architecture can also tackle the over-smoothing problem. Many GNNs, including our proposed models, are compatible with them. We will not precisely introduce or compare them.

In this paper, we refine and propose Graph Diffusion Networks (GDNs). Then we propose Adaptive Graph Diffusion Neural Networks (AGDNs) with generalized graph diffusion associated with two learnable and scalable mechanisms of weighting coefficients: Hop-wise Attention (HA) and Hop-wise Convolution (HC). We show a natural evolution path of GNNs. From MPNNs to GDNs, the receptive field is enlarged without adding extra transformations or decoupling the model architecture. The multi-layer graph diffusion in different feature spaces can contribute to model capacity. From GDNs to AGDNs, a more generalized and flexible paradigm of graph diffusion can bring better performance. HA induces hop-wise and node-wise weighting coefficients. HC can directly learn the hop-wise and channel-wise weighting coefficients. Following the historical development of GNNs, from elegant but inefficient spectral methods to intuitive and effective spatial methods, we generalize the graph diffusion to be more spatial with a loss of spectral analyzability. 

We conduct experiments on diverse Open Graph Benchmark (OGB) \cite{hu2020open} datasets with semi-supervised node classification and link prediction tasks \cite{hu2020open}. The results show that our proposed methods significantly outperform popular GNNs on both tasks. AGDNs achieve \textbf{top-1} performance on the ogbn-arxiv (Accuracy of 76.37±0.11\%), ogbn-proteins (ROC-AUC of 88.65±0.13\%) and ogbl-ddi (Hits@20 of 95.38±0.94\%) datasets and \textbf{top-3} performance on the ogbl-citation2 (MRR of 85.49±0.29\%) dataset. AGDNs also achieve the SOTA performance among models without using labels as input on the ogbn-products dataset. AGDNs outperform the state-of-the-art (SOTA) RevGNNs with much less complexity and runtime and achieve comparable results to SOTA SEAL with much less runtime on large graphs. In our ablation study on all datasets, AGDNs can significantly outperform the associated GATs. Furthermore, the experiments of MPNNs and AGDNs with different model depths demonstrate that AGDNs can effectively mitigate the over-smoothing effect.

Our main contributions are 1). We incorporate multi-layer generalized graph diffusion into GNNs with moderate complexity and runtime; 2). We propose two learnable and scalable mechanisms for adaptively capturing multi-hop information. 3). We achieve new SOTA performance on diverse, challenging OGB datasets. We outperform complicated RevGNNs with much less complexity and runtime on large datasets.

\section{Related works}
\subsection{Residual GNNs}
By employing residual connections, residual GNNs simultaneously alleviate over-fitting and over-smoothing problems. The main concerns of these methods are connection design and memory optimizations. Jumping-Knowledge Network (JKNet) \cite{xu2018representation} introduces jumping knowledge connection and adaptively sums intermediate representations. GCN with initial residual and identity mapping (GCNII) \cite{chen2020simple} combines two classes of residual connections. DeepGCN \cite{li2019deepgcns} introduces dense and dilated connections from CNNs. DeeperGCN \cite{li2020deepergcn} unifies message aggregation operations with differentiable generalized aggregation functions. It further proposes a novel normalization layer and pre-activation residual GNNs. Reversible GNNs (RevGNNs) \cite{li2021training} reduce memory footprints by incorporating reversible connections and grouped convolutions. With deep and over-parameterized GNN architecture, RevGNNs can achieve SOTA results on several datasets. However, as a cost of reducing memory footprints, the runtime of deep RevGNNs layers is even longer. This paper will present deep GNNs with shallow feature transformations that outperform SOTA RevGNNs with much less complexity and runtime.

\subsection{Graph diffusion}

We describe the graph convolution as a matrix multiplication between the transition matrix and node feature/representation matrix. Then the graph diffusion \cite{klicpera2018predict,klicpera2019diffusion,page1999pagerank,kondor2002diffusion} replaces the transition matrix with a diffusion matrix, which is a linear combination of powers of the transition matrix with weighting coefficients normalized along hops.
The weighting coefficients are essential to balance the importance of shallow and deep information. The Personalized PageRank (PPR) \cite{page1999pagerank,klicpera2018predict} and the Heat Kernel (HK) \cite{kondor2002diffusion,xu2020graph} are two popular predefined weighting coefficients. They both follow the prior that more distant neighboring nodes have fewer influences than near neighboring ones. The weighting coefficients can also be analogously trainable parameters using label propagation \cite{berberidis2018adaptive, chen2013adaptive}. Attention walk \cite{abu2018watch} jointly optimizes the node embeddings and weighting coefficients. 

\subsubsection{Decoupled GNNs}
To perform deep graph convolutions with shallow feature transformations, decoupled GNNs decouple graph convolution operators and feature transformations. The integration of multi-hop information in several decoupled GNNs can be considered the special cases \cite{wu2019simplifying, klicpera2018predict, zhu2021simple} or generalized cases \cite{liu2020towards} of the graph diffusion. The Approximated Personalized Propagation of Neural Predictions (APPNP) \cite{klicpera2018predict} utilizes the predefined weights exponentially decaying with hops to integrate multi-hop information. The Deep and Adaptive Graph Neural Network (DAGNN) \cite{liu2020towards} incorporates learnable weights. They both perform graph convolutions after feature transformations. On the contrary, other decoupled GNNs \cite{wu2019simplifying, zhu2021simple, rossi2020sign} perform graph convolutions before feature transformations. The Simplified Graph Convolution network (SGC) \cite{wu2019simplifying} utilizes the last representation and then classifies with a linear layer. SGC is significantly affected by the over-smoothing problem. As an early GNN model, Diffusion Convolution Neural Network (DCNN) \cite{atwood2016diffusion} can also be considered a decoupled GNN using the graph diffusion operator with a learnable convolution kernel. DCNN directly performs graph diffusion with learnable hop-wise and channel-wise weighting coefficients (convolution kernel) based on the input node feature matrix and applies a linear layer. Although DCNN finalizes the graph diffusion during model training, it calculates and stores high powers of the transition matrix, which limits its scalability. Based on SGC, the Simple Spectral Graph Convolution network (S$^2$GC) sums multi-hop features. The Scalable Inception Graph Neural Network (SIGN) \cite{rossi2020sign} encodes and concatenates multi-hop representations. However, as a trade-off between efficiency and accuracy, the simple decoupled architecture limits model capacity because all graph convolution operators are restricted in the same feature space. There are no intermediate transformations between graph convolution operators. 

In this paper, we tend to directly replace the graph convolution operator with a graph diffusion operator in each MPNN layer without decoupling the model architecture. We must consider whether to calculate an explicit diffusion matrix during actual implementation.
\subsubsection{Explicit diffusion matrix: Graph Diffusion Convolution}
As a preprocessing method to augment input data, the Graph Diffusion Convolution (GDC) \cite{klicpera2019diffusion} calculates an explicit diffusion matrix. Although GDC controls the sparsity of the final diffusion matrix, the intermediate explicit high-power transition matrices are still maintained, which limits its scalability. There are approximation methods for PPR and HK to alleviate this problem with a loss of information. Moreover, the weighting coefficients are predefined and identical for all nodes, channels, and layers, which may limit the model performance. GDC cannot be calculated in preprocessing for a learnable transition matrix. In addition, GDC cannot improve link prediction performance.

\subsubsection{Implicit diffusion matrix: Graph Diffusion Networks}
The Graph Diffusion Networks (GDNs) perform implicit graph diffusion with right-to-left matrix multiplication starting from the node feature or representation matrix in each layer. First, GDNs sequentially calculate the multi-hop representation matrices. Then, they integrate these matrices without maintaining a high-dimensional diffusion matrix. Such a paradigm is more memory-efficient because high-order aggregated feature matrices are much smaller than high-power transition matrices. Another critical advantage over GDC is that the weighting coefficients in GDNs can be layer-wise. GDNs multiply the receptive field of base MPNNs with the diffusion depth without increasing feature transformations or changing other central model architecture. 
Thus, GDNs can inherit many essential characteristics from MPNNs, including non-linear hierarchies, attention mechanisms, and residual connections. In contrast, decoupled GNNs lose most of them. Based on an MPNN layer, we can generate its associated GDN layer. Topology Adaptive Graph Convolutional Networks (TAGCNs) \cite{du2017topology} are generated based on GCN and sum multi-hop representations after transformations using uniform weighting coefficients. TAGCNs utilize a generalized version of the graph diffusion since multiple linear transformations are applied to multi-hop features. These additional transformations bring redundant complexity. MAGNA \cite{wang2020direct} is generated based on GAT using PPR weighting coefficients. MAGNA further includes residual connections and feed-forward networks for each layer. In addition, GDNs using PPR or HK coefficients with a predefined transition matrix are equivalent to GDC.
However, previous works have not precisely studied the unified framework of GDNs. In this paper, we refine the unified framework of GDNs and extend this framework with generalized graph diffusion.  

\subsubsection{Diffusion-like GNNs}

MixHop \cite{abu2019mixhop} and N-GCN \cite{abu2020n} perform diffusion-like operation in each layer, however, they transform node representations with non-linear activations for each hop. Such methods more likely stack MPNN layers into groups by concatenating their outputs without effectively controlling model complexity. They are hard to optimize and have redundant hyperparameters with a complicated tuning strategy, including selecting which hops to use in each layer. Our proposed GDNs and AGDNs follow a more efficient paradigm of graph diffusion without redundant feature transformations or non-linear activations, and AGDNs can adaptively learn the importance of each hop instead of manual selection.

\subsection{Tricks on GNNs}
The Bag of Tricks on GNNs (BoT)\cite{wang2021bag} includes some critical tricks of GNNs applied in many SOTA models on the OGB leaderboard. Firstly, the masked node labels, which are zeros other than sampled training nodes, are used as model input ("label as input"). Secondly, BoT conducts the additional iterative feed-forward passes in each epoch with the predicted soft labels filling up masked zero labels ("label reuse"). Thirdly, BoT proposes a more robust loss function ("Loge loss"). Finally, BoT proposes to adjust the GAT adjacency closer to the GCN adjacency ("norm.adj."). Self-Knowledge Distillation (self-KD) \cite{zhang2019your} is another common trick on the OGB leaderboard. Graph Information Aided Node feature exTraction with XR-Transformers (GIANT-XRT) \cite{chien2021node} trains more informative node embedding using raw text data.

\subsection{GNNs for link prediction}
There exist two main GNN methods applied to link prediction. The first methods follow an encoder-decoder framework \cite{kipf2016variational, hu2020open}, in which GNNs act as the encoder to output node representations. A simple predictor receiving pairs of node representations makes final predictions. Pairwise Learning on Neural Link Prediction (PLNLP) proposes to utilize pairwise AUC loss to improve the quality of ranking metrics. The second method, SEAL \cite{zhang2021labeling, zhang2018link} applies a GNN on an enclosing subgraph sampled from each pair of nodes and directly output predictions. SEAL also includes an additional labeling trick to enhance structural information. SEAL achieves the SOTA performance of GNNs on many link prediction datasets. However, SEAL requires extensive runtime, which limits its application, especially on large graphs. This paper will show that with the encoder-decoder framework, AGDNs can outperform other encoder-decoder GNNs and even approach or outperform SEAL.

\section{Preliminaries}
\subsection{Notations}
Suppose $\cal G=(\cal V, \cal E)$ be a given graph with node set $\cal V$ and edge set $\cal E$. We denote the number of nodes with $N=|{\cal V}|$, the number of edges with $E=|\cal E|$, and the adjacency matrix with ${\boldsymbol A}\in \mathbb R^{N \times N}$. The normalized adjacency matrix is denoted with $\overline{\boldsymbol A}\in \mathbb R^{N\times N}$.
Considering the common stacking architecture of GNNs, we denote the initial node feature matrix with $\boldsymbol X \in {\mathbb R}^{N\times d^{(0)}}$ and the initial edge feature matrix with $\boldsymbol X^{E}$. For the $l$-th GNN layer, we denote its input node representation with $\boldsymbol H^{(l-1)}\in {\mathbb R}^{N\times d^{(l-1)}}$ and its output node representation with ${\boldsymbol H}^{(l)}\in {\mathbb R}^{N\times d^{(l)}}$. We also denote node $i$'s input representation vector with $\boldsymbol h_i^{(l-1)}$ and its output representation vector with $\boldsymbol h_i^{(l)}$. We denote the attention query vector of GAT with $\boldsymbol a^{(l)}\in {\mathbb R}^{2\times d^{(l)}}$ and the hop-wise attention query vector of AGDN-HA with $\boldsymbol a_{hw}^{(l)}\in {\mathbb R}^{2\times d^{(l)}}$.

\begin{table}[!hbt]
\small
\caption{Symbol definitions.}\label{tab: symbol definitions}
\begin{center}
  \setlength{\tabcolsep}{1.mm}{
    \begin{tabular}{c|c|c|c}
    \toprule
      \textbf{Symbols}  & \textbf{Definitions} & \textbf{Symbols}  & \textbf{Definitions} \\
      \midrule
      $\mathcal G$ & Graph & $K$ & Diffusion depth \\
      $\mathcal{V}$ & Node set & $N$ & Number of nodes\\
      $\mathcal{E}$ & Edge set &  $E$ & Number of edges \\
      $\boldsymbol A$ & Adjacency matrix & $\overline{\boldsymbol {A}}$ & transition matrix  \\
      $\boldsymbol X$ & Node feature matrix & $\boldsymbol x_i$ & Node feature vector \\
      $\boldsymbol X^{E}$ & Edge feature matrix & $\boldsymbol x^{E}_{(i,j)}$ & Edge feature vector \\
      $\boldsymbol H^{(l)}$ & Node representation matrix & $\boldsymbol h^{(l)}_i$ & Node representation vector \\
      $d^{(0)}$ & Node feature dimension & $d^{(l)}$ & $l$-th layer's output dimension \\
      $d^{E}$ & Edge feature dimension &$\mathcal{N}_i$ & Neighborhood set of node $i$\\

    \bottomrule
    \end{tabular}
    }
\end{center}
    
\end{table}

\subsection{Tasks}
\subsubsection{Semi-supervised node classification}
Given a graph $\mathcal G$, the node set can be further split into a labeled node set $\mathcal L$ and an unlabeled node set $\mathcal U$. The semi-supervised node classification task is to predict unlabeled nodes' labels. The unlabeled nodes' feature and related edges are not accessible at inference for inductive tasks. Nevertheless, they are accessible for transductive tasks. For inductive tasks, unlabeled nodes' all information cannot be accessed during training. Whether GDNs or AGDNs are applicable on inductive datasets ultimately depends on its MPNN base model.

\subsubsection{Link prediction}
Given a graph $\mathcal G$ and a pair of nodes, the link prediction task is to predict whether an edge exists between two nodes. The edges in a dataset are split into training, validation, and test edges. OGB rules make the training edges accessible during model training and inference. The other edges can only be accessed to compute evaluation metrics.

\subsection{Message Passing Neural Networks}
Let us review the main architecture of MPNNs. This paper focuses on the essential step in MPNNs, neighbor aggregation (message passing). We also consider the following combination with node self feature matrix (usually in the form of residual linear connection). We omit the optional readout operation (only necessary for the graph prediction). 
The neighbor aggregation with a residual linear connection can be described as a matrix multiplication between weighted adjacency $\overline{\boldsymbol A}$ (transition matrix) and node feature/representation matrix $\boldsymbol H$:
\begin{equation}
    {\boldsymbol H}^{(l)}=\boldsymbol {\overline A}^{(l)}{\boldsymbol H}^{(l-1)}{\boldsymbol W}^{(l)}+\boldsymbol{H}^{(l-1)}\boldsymbol{W}^{(l),r},
\end{equation}
where $\boldsymbol W\in \mathbb R^{d^{(l-1)}\times d^{(l)}}$ refers to the linear transformation in the $l$-th layer, and $\boldsymbol{W}^{(l),r}\in \mathbb R^{d^{(l-1)}\times d^{(l)}}$ refers to its residual linear transformation. A complete MPNN model stacks several MPNN layers with intermediate activations.

\subsubsection{Transition matrix}
The weighted adjacency or transition matrix is used in graph convolution (1-power) and graph diffusion (multiple powers). There are several common transition matrices: row-stochastic adjacency ${\overline {\boldsymbol A}}_{row}={\boldsymbol D}^{-1}{\boldsymbol A}$, column-stochastic adjacency $\overline{\boldsymbol A}_{col}={\boldsymbol A}{\boldsymbol D}^{-1}$ and the symmetric normalized adjacency $\overline{\boldsymbol A}_{sym}={\boldsymbol D}^{-1/2}{\boldsymbol A}{\boldsymbol D}^{-1/2}$, where ${\boldsymbol D}$ is the degree matrix (${D}_{ii}=\Sigma^N_{j=1}{ {A}_{ij}}$). Different transition matrices are associated with different base MPNN models. GraphSAGE utilizes $\overline{\boldsymbol A}_{row}$ as its weighted adjacency. GCN utilizes $\overline{\boldsymbol A}_{sym}$ modified with adding self-loops. GAT utilizes learnable row-stochastic adjacency, derived from the learnable attention mechanism, which depends on the attributes of source and destination nodes. We can view this adjacency as a learnable row-stochastic adjacency by reviewing the computation of GAT adjacency:
\begin{equation}
\begin{split}
e_{ij}&=\mathop{\rm LeakyReLU}([\boldsymbol h_i\boldsymbol W\lvert\rvert \boldsymbol h_j\boldsymbol W]\cdot \boldsymbol a)\\
      &=\mathop{\rm LeakyReLU}([\boldsymbol h_i\boldsymbol W]\cdot\boldsymbol a^{dst}+[\boldsymbol h_j\boldsymbol W]\cdot\boldsymbol a^{src}),
\end{split}
\end{equation}
\begin{equation}
\overline {\mathcal A}_{ij} = \frac{\mathop{\rm exp}(e_{ij})}{\sum_{k\in \mathcal N_i}\mathop{\rm exp}(e_{ik})},
\label{eq: GAT softmax}
\end{equation}
where $\boldsymbol h_i$ is the input representation vector of node $i$, $\boldsymbol W$ is the transformation matrix. The query vector $\boldsymbol a$ can be split into source query vector $\boldsymbol a^{src}$ and destination query vector $\boldsymbol a^{dst}$.

For graphs with edge attributes $\boldsymbol x_{ij}^E$, BoT \cite{wang2021bag} proposes to incorporate edge attributes in attention coefficients:
\begin{equation}
    e_{ij}=\mathop{\rm LeakyReLU}([\boldsymbol h_i\boldsymbol W\lvert\rvert \boldsymbol h_j\boldsymbol W\lvert\rvert \boldsymbol x_{ij}^{E} \boldsymbol W^{E}]\cdot \boldsymbol a),
\end{equation}
where $\boldsymbol W^{E}$ is edge feature transformation matrix.

Then, we can define the unnormalized GAT adjacency $\boldsymbol {\mathcal A}: \mathcal A_{ij}=\mathop{\rm exp}(e_{ij})$ and its diagonal in-degree and out-degree matrices are defined as $\boldsymbol {\mathcal D}_{row}$ and $\boldsymbol {\mathcal D}_{col}$, with respectively row summation and column summation as diagonal entries. The GAT adjacency may be expressed as ${\overline{\boldsymbol{\mathcal A}}}=\boldsymbol {\mathcal D}_{row}^{-1}\boldsymbol {\mathcal A}$.

\section{Proposed methods}

In this section, we firstly introduce the frameworks of GDNs and AGDNs. Then, we propose a pseudo-symmetric variant of the GAT transition matrix. Thirdly, we propose two adaptive mechanisms for calculating hop weighting matrices. 

\subsection{Graph Diffusion Networks}
Several existing GNNs using multi-hop information in each layer can be considered GDNs. However, there is a lack of a unified framework for GDNs. For a GDN model, we also stack multiple GDN layers to perform multi-layer graph diffusion. We formulate a GDN layer with the diffusion depth $K$ as below:
\begin{equation}
    \tilde{\boldsymbol H}^{(l,0)} = \boldsymbol H^{(l-1)}\boldsymbol W^{(l)},
\end{equation}
\begin{equation}
    \tilde{\boldsymbol H}^{(l,k)} = \overline{\boldsymbol A}\tilde{\boldsymbol H}^{(l,k-1)},
\end{equation}
\begin{equation}
    {\boldsymbol H}^{(l)}=\sum^{K}_{k=0}\theta^{(l,k)}\tilde{\boldsymbol H}^{(l,k)}+\boldsymbol{H}^{(l-1)}\boldsymbol{W}^{(l),r},
\end{equation}
where $\{\theta^{(k)}\}_{k\in\{0,1,...,K\}}$ is the set of normalized weighting coefficients and $K$ is the diffusion depth. Note that we iteratively calculate the multi-hop representations in a right-to-left matrix multiplication, instead of the explicit multiple powers of the transition matrix ($\overline{\boldsymbol A}^{(l)}, (\overline{\boldsymbol A}^{(l)})^2,..,(\overline{\boldsymbol A}^{(l)})^K$).

\subsection{Adaptive Graph Diffusion Networks}

\begin{figure}[tbp]
    \centering
    \includegraphics[width=2.5in, keepaspectratio]{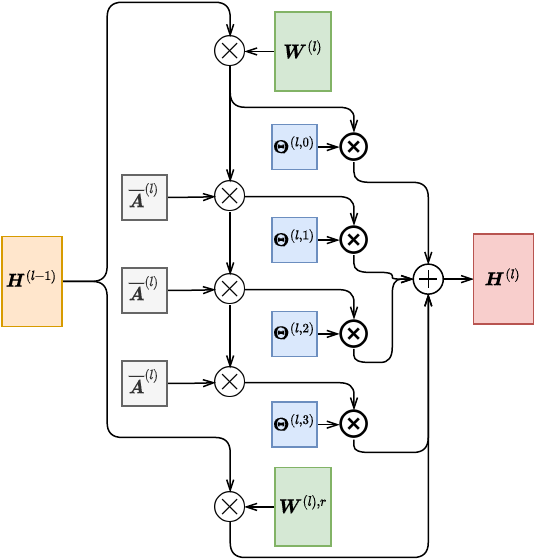}
    \caption{\textbf{AGDN layer Architecture}: The operator $\otimes$ represents matrix multiplication, the \textbf{bold} operator $\boldsymbol\otimes$ represents element-wise multiplication, and the operator $\oplus$ represents summation. The left and right multiplication correspond to the relative position of the multiplicand to the operator $\otimes$.}
    \label{fig:AGDN layer}
\end{figure}

The weighting coefficients in GDC and GDNs are hop-wise. In this paper, we generalize the graph diffusion and make weighting coefficients further node-wise or channel-wise. We suppose that different nodes or channels may require different hop weighting coefficients. We define an AGDN layer as below:
\begin{equation}
    \tilde{\boldsymbol H}^{(l,0)} = \boldsymbol H^{(l-1)}\boldsymbol W^{(l)},
\end{equation}
\begin{equation}
    \tilde{\boldsymbol H}^{(l,k)} = \overline{\boldsymbol A}\tilde{\boldsymbol H}^{(l,k-1)},
\end{equation}
\begin{equation}
    {\boldsymbol H}^{(l)}=\sum^{K}_{k=0}{\boldsymbol \Theta}^{(k)}\otimes\tilde{\boldsymbol H}^{(l,k)}+\boldsymbol{H}^{(l-1)}\boldsymbol{W}^{(l),r},
\label{eq: AGDN}
\end{equation}
where $\otimes$ refers to the element-wise multiplication and ${\boldsymbol \Theta}^{(k)}$ in $\mathbb R^{N\times d^{(l)}}$ is a generalized weighting matrix. GDNs are special cases of AGDNs when all elements of ${\boldsymbol \Theta}^{(k)}$ are the same. This description in the form of matrices is for global comparison with MPNNs and GDNs.

In detail, we also describe the AGDN layer from a node viewpoint, which matches the actual implementation. We perform sequential graph convolution and sum multi-hop representations with hop-wise (node-wise or channel-wise) weighting coefficients. The generalized graph diffusion at $l$-th layer for node $i$ is described as below:
\begin{equation}
    \tilde{\boldsymbol h}^{(l,0)}_i=\boldsymbol h^{(l-1)}_i\boldsymbol W^{(l)},
\end{equation}
\begin{equation}
    \tilde{\boldsymbol h}^{(l,k)}_i=\sum_{j\in \mathcal{N}_i}\overline{A}_{ij}\tilde{\boldsymbol h}^{(l,k-1)}_j,
\end{equation}
\begin{equation}
    h^{(l)}_{ic}=\sum_{k=0}^{K}\theta_{ikc}\tilde{h}^{(l,k)}_{ic}+\sum_{c'=1}^{d^{(l-1)}}h^{(l-1)}_{ic'}W^{(l),r}_{c'c},
\end{equation}
where $\tilde{\boldsymbol h}^{(l,k)}_i$ is the $k$-hop intermediate representation vector of node $i$ and $\theta_{ikc}$ can be viewed extracted from a 3-dimensional (node-wise, hop-wise and channel-wise) tensor $\boldsymbol \Theta$. The previous weighting matrix $\boldsymbol \Theta^{(k)}$ can be extracted from this tensor by selecting the $k$-hop. In some cases, to explicitly enhance the position (hop) information, we can add intermediate multi-hop representation vectors with learnable Positional Embedding (PE) row vectors $\{\boldsymbol p^{(0)}, \boldsymbol p^{(1)}, ..., \boldsymbol p^{(K)}\}$ in $\mathbb R^{d^{(l)}\times 1}$. We omit this trick since PE marginally improves model performance on certain datasets empirically.

There exists a trade-off between generalization and spectral analyzability. The eigenvectors of the generalized graph diffusion matrix are generally different from the original transition matrix. From another perspective, AGDNs do not follow the characteristics or limitations of previous diffusion-based methods. Instead, our generalized weighting coefficients can be adaptive across hops and nodes or channels, reflected in the following proposed Hop-wise Attention and Hop-wise Convolution. In addition, we can naturally assign layer-wise weighting coefficients.


\subsection{Transition matrix}

We can find that GAT is more concerned about destination nodes' in-degrees. The symmetric normalized adjacency has proven more effective on specific datasets \cite{kipf2016semi}. It is reasonable to give a variant of GAT adjacency to leverage both source nodes' out-degrees and destination nodes' in-degrees. Thus, motivated by the form of popular symmetric normalized adjacency, we propose a pseudo-symmetric normalized GAT adjacency:
\begin{equation}
\overline{\boldsymbol{\mathcal{A}}}_{sym}=\boldsymbol{\mathcal D}_{row}^{-\frac{1}{2}}\boldsymbol{\mathcal A}\boldsymbol{\mathcal D}_{col}^{-\frac{1}{2}}.
\end{equation}

In BoT, another version of pseudo-symmetric normalized GAT adjacency is proposed (denoted with "norm.adj.")  \cite{wang2021bag}:
\begin{equation}
\overline{\boldsymbol{\mathcal A}}_{adj}=\boldsymbol D^{\frac{1}{2}}\boldsymbol{\mathcal D}_{row}^{-1}\boldsymbol{\mathcal A}{\boldsymbol D}^{-\frac{1}{2}},
\end{equation}
where $adj$ refers to "adjustment" since we can view this adjacency as the GAT adjacency adjusted to GCN adjacency. Note that $\boldsymbol {\mathcal A}$, $\overline{\boldsymbol{\mathcal{A}}}_{sym}$ and $\overline{\boldsymbol{\mathcal A}}_{adj}$ are pseudo-symmetric since they are guaranteed to be symmetric if and only if $e_{ij}=e_{ji}, \forall i,j\in \mathcal N$, when $\boldsymbol a^{src}=\boldsymbol a^{dst}$. As a special case when we set query vectors to zeros, then both ${\overline{\boldsymbol{\mathcal A}}}_{sym}$ and ${\overline{\boldsymbol{\mathcal A}}}_{adj}$ become the standard symmetric normalized adjacency. This characteristic connects GAT and GCN.

\subsection{Weighting ceofficients}
In this subsection, to simplify the discussion, we omit the subscript $(l)$ and collect all weighting coefficients $\{\theta_{ikc}\}$ into a unified weighting tensor with $\boldsymbol \Theta \in \mathbb R^{N\times (K+1)\times d}$. We can extract ${\boldsymbol \Theta}^{(k)}={\boldsymbol \Theta}_{:k:}$, where $:$ in the subscript refers to extracting all channels in this dimension. We denote the subscripts for nodes, hops, and channels with $i$, $k$, and $c$. We aim to design 'adaptive' and efficient ways of calculating weighting tensor, which should be variable for nodes or feature channels. It is hard to directly define a unified weighting tensor adaptive for both nodes and feature channels, which results in enormous additional complexity. We propose two efficient ways in the following paragraphs: hop-wise attention (HA) and hop-wise convolution (HC). We denote AGDN variants with AGDN-mean, AGDN-HA, and AGDN-HC, using naive fixed weights, hop-wise attention weights, and hop-wise convolution weights. As a particular case of AGDNs, AGDN-mean can be considered a representative example of GDNs.

\begin{figure}[tbp]
\centering
\includegraphics[scale=0.5]{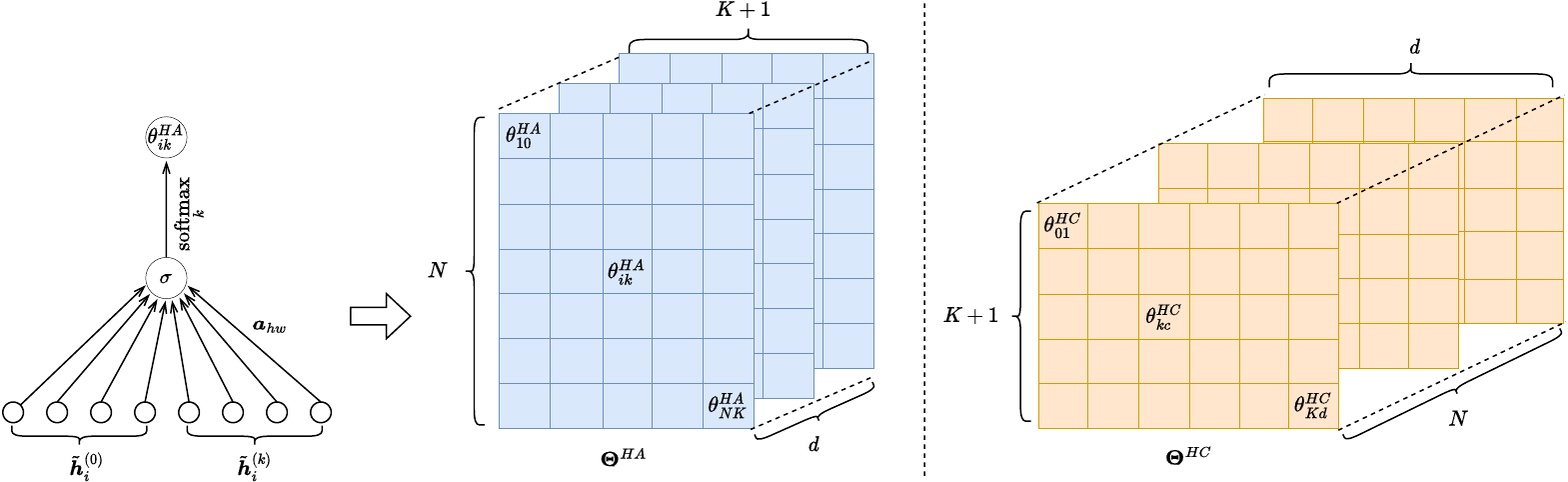}
\caption{\textbf{Left}: The hop-wise attention is parameterized by $\boldsymbol a_{hw}$, with a LeakyReLU activation function $\sigma$ and normalized along hops with softmax function; The associated weighting tensor $\boldsymbol \Theta$ can be derived from the $N\times (K+1)$ matrix $\boldsymbol \Theta^{HA}$, by repeating $d$ times along the third dimension. \textbf{Right}: The weighting tensor of hop-wise convolution $\boldsymbol \Theta$ is directly derived from $(K+1)\times d$ kernel matrix $\boldsymbol \Theta^{HC}$ by repeating $N$ times along the first dimension.}

\label{fig: HA and HC}

\end{figure}

\paragraph{Hop-wise Attention}
We suppose that, in many cases, the computation of graph diffusion should be adaptive for different nodes but identical for different feature channels, which manifests as the unified weighting tensor normalized along hops $\sum_{k=0}^{K}{\theta}_{i,k,c}=1, \forall i,c$
and identical along channels ${\theta}_{i,k,c}={\theta}_{i,k,1}, \forall c\in\{1,2,...,d\}$. Then we can simplify this tensor into a 2-dimensional weighting matrix $\boldsymbol \Theta^{HA}=[\theta^{HA}_{ik}]$ in $\mathbb R^{N\times (K+1)}$, by ignoring the last subscript $c$. $\boldsymbol \Theta$ can be recovered by adding the third dimension and repeating $\boldsymbol \Theta^{HA}$ for $d$ times in the third dimension. It is still not efficient to define a naive learnable weighting matrix in $\mathbb R^{N\times (K+1)}$, which results in redundant complexity and violates the possible inductive setting. Inspired by the attention mechanism in GAT, we propose Hop-wise Attention (HA) using a learnable query vector $\boldsymbol a_{hw}$ in $\mathbb R^{2d}$ to induce the expected weighting matrix. We need to learn just $2d$ parameters. Firstly, we calculate $\omega$:
\begin{equation}
    \omega_{ik} = \left[\tilde{\boldsymbol h}^{(l,0)}_{i}\left\lvert\right\rvert\tilde{\boldsymbol h}^{(l,k)}_{i}\right]\cdot \boldsymbol a_{hw},
\end{equation}
where $\cdot$ represents inner product, $\lvert\rvert$ represents the concatenation operation and $k$ represents the $k$-hop representation.
Then, as shown in the left part of Figure \ref{fig: HA and HC}, the hop-wise attention scores are calculated as below:
    
    \begin{equation}
    {\theta}_{ik}^{HA}=\frac {{\rm exp}\left({\sigma}\left(\omega_{ik}\right)\right)}{\sum _{k=0}^{K} {{\rm exp}\left({\sigma}\left(\omega_{ik}\right)\right)}},
    \end{equation}
where $\sigma$ is an activation function (usually LeakyReLU).

\paragraph{Hop-wise Convolution}

We consider another simple strategy for integrating multi-hop representations, which performs Hop-wise Convolution (HC). This time, we suppose that, in certain cases, the graph diffusion should be adaptive for different channels. Thus, we directly define learnable weighting tensor, which is identical for all nodes ${\theta}_{i,k,c}={\theta}_{1,k,c},\forall i\in \{1,2,...,N\}$. Then we simplify this tensor into a 2-dimensional convolution kernel matrix $\boldsymbol \Theta^{HC}=[\theta^{HC}_{kc}]$ in $\mathbb R^{(K+1)\times d}$, by ignoring the first subscript $i$. The complete weighting tensor $\boldsymbol \Theta$ can be recovered by adding the first dimension and repeating $\boldsymbol \Theta^{HC}$ for $N$ times in the first dimension, as shown in Figure \ref{fig: HA and HC}. We need to learn $(K+1)\times d$ parameters. Note that we do not require that the tensor is normalized along any dimensions. For each feature channel $c\in \{1,2,...,d\}$, we conduct individual hop-wise convolution with the associated kernel vector in $\mathbb R^{K+1}$. HC is in the same form as DCNN. However, HC is based on more memory-efficient graph diffusion and calculated with different convolution kernels in different layers.




\subsection{Complexity}
In complexity analysis, we omit the dimension change between a layer's input and output. The extra time complexity of an AGDN layer over its base MPNN layer comes from $K$-hop aggregations $O(KEd)$ (by default, we perform feature transformation before aggregation), element-wise multiplication with weighting matrices $O(KNd)$, and hop-wise attention computation $O(KNd)$ if used. Then the extra time complexity of an AGDN layer is $O(KEd+KNd)$. Under the realistic assumption of $E\ll N$, this extra time complexity becomes $O(KEd)$.
The extra space complexity of an AGDN layer is $O(KNd)$.

\section{Model analysis}
We generally lose the elegant spectral analyzability since generalized weighting coefficients can easily change the eigenvectors. However, this also implies that AGDNs are more flexible in the spectral domain. 

For AGDN-HC, the weighting coefficients are identical for all nodes and do not change the eigenvectors. Thus, we can give a preliminary spectral analysis. We will demonstrate that, even without changing eigenvectors, AGDN-HC is still flexible in the spectral domain with a considerable diffusion depth. First, given a feature channel $c$, we simplify the form of AGDN-HC with an eigendecomposition of the transition matrix $\overline{\boldsymbol A}=\boldsymbol U^{-1} \boldsymbol \Lambda \boldsymbol U$:
\begin{equation}
\begin{split}
    \boldsymbol{S}_{c}&=\sum_{k=0}^{K}\theta_{kc} \overline{\boldsymbol{A}}^{k}\\
                &=\sum_{k=0}^{K}\theta_{kc}\left(\boldsymbol U^{-1}\boldsymbol{\Lambda}\boldsymbol{U}\right)^{k}\\
                &=\boldsymbol{U}^{-1}\left(\sum_{k=0}^{K}\theta_{kc} \boldsymbol{\Lambda}^{k}\right)\boldsymbol{U},
\end{split}
\end{equation}
where the row vectors of $\boldsymbol U$ refer to the eigenvectors of $\overline{\boldsymbol A}$ and $\boldsymbol \Lambda$ is a diagonal matrix whose entries $\{\lambda_1,\lambda_2,...,\lambda_N\}$ are eigenvalues of $\overline{\boldsymbol A}$. The eigenvalues of the transition matrix are bounded by 1 ($\lambda_i\in[-1,1],\forall i$) \cite{ng2001spectral}.
For the $i$-th eigenvalue $\lambda_i$ of the transition matrix, the associated eigenvalue $\lambda'_i$ of the diffusion matrix is:
\begin{equation}
    \lambda'_i = \sum_{k=0}^{K}\theta_{k,c}\lambda^{k}_i.
\end{equation}
This relation can be described as an $K$-order polynomial function of $\lambda_i$. With the order increasing, this function becomes more flexible and can approximate more functions. 

\section{Experiments}
In this section, we conduct experiments on three OGB node classification datasets and three OGB link prediction datasets. Our proposed AGDNs outperform common MPNNs and SOTA RevGNNs with less complexity and runtime for the semi-supervised node classification datasets. AGDNs outperform other GNN models in link prediction tasks using the same encoder-decoder framework. AGDNs approach SOTA SEAL with much less runtime. AGDNs achieve new SOTA performance on the ogbn-arxiv, ogbn-proteins, and ogbl-ddi datasets. We train all AGDN models on a single V100 card with 16Gb memory.

\begin{table}[htb]
    \caption{Dataset statistics}
\begin{center}
    \begin{tabular}{c|ccc}
    \toprule[1pt]
         \textbf{Datasets} & \textbf{\#Nodes} & \textbf{\#Edges} & \textbf{Metrics}  \\
         \midrule
         ogbn-arxiv & 169,343 & 1,166,243 & Accuracy \\
         ogbn-proteins & 132,534 & 39,561,252 & ROC-AUC \\
         ogbn-products & 2,449,029 & 61,859,140 & Accuracy \\
         \midrule
         ogbl-ppa & 576,289 & 30,326,273 & Hits@100 \\
         ogbl-ddi & 4,267 & 1,334,889 & Hits@20 \\
         ogbl-citation2 & 2,927,963 & 30,561,187 & MRR \\
    \bottomrule[1pt]
    \end{tabular}
\end{center}
    \label{tab:dataset statistics}
\end{table}

\paragraph{Datasets}
We utilize three OGB semi-supervised node classification datasets (ogbn-arxiv, ogbn-proteins and ogbn-products) and three OGB link prediction datasets (ogbl-ppa, ogbl-ddi and ogbl-citation2). We summarize the detailed statistics of these datasets in Table \ref{tab:dataset statistics}. ogbn-arxiv is a citation network between all Computer Science (CS) arXiv papers, whose data split is based on the publication dates of the papers. ogbn-proteins is a graph between proteins with multi-dimensional edge weights indicating different types of biologically meaningful associations. Its data split is based on the associated species of the proteins. ogbn-products is a co-purchasing network between Amazon products, whose data split is based on the sales ranking. 
ogbl-ppa is a graph between proteins from 58 species with similar edges to ogbn-proteins, whose edges measured by high-throughput technology are used as training edges, and other edges measured by low-throughput technology are used as validation and testing edges. ogbl-ddi is a drug-drug interaction network with each edge indicating the joint effect of taking the two drugs together. It has data split based on what proteins those drugs target in the body. ogbl-citation2 is a citations graph between a subset of papers from MAG. Its data is split by selecting the most recent papers as source nodes and randomly selecting destination nodes for training/validation/testing sets.

\paragraph{Global settings}
We conduct all experiments on a single Nvidia Tesla V100 with 16 Gb GPU memory. We evaluate our proposed models with 10 runs, fixing random seed 0-9, and report means and standard deviations. Except on the ogbn-products and ogbl-citation2 datasets (evaluated on CPU), we conduct both training and inference of all AGDN models on the same GPU card. All final test scores are from the best model selected based on validation scores. In the tables of this paper, we highlight the results of AGDN with \underline{underlined fonts} and the best results with \textbf{bold fonts}. We utilize AGDN-HC on the ogbn-proteins dataset and AGDN-HA for all other datasets. The unavailable results are indicated by –.

\subsection{Task 1: Node Classification}

\paragraph{Baselines}
Several representative GNNs and SOTA GNNs are selected as baselines. For semi-supervised node classification, we utilize GCN \cite{kipf2016semi}, GraphSAGE \cite{hamilton2017inductive}, GAT \cite{velivckovic2017graph}, MixHop \cite{abu2019mixhop}, JKNet \cite{xu2018representation}, DeeperGCN \cite{li2020deepergcn}, GCNII \cite{chen2020simple}, DAGNN \cite{liu2020towards}, MAGNA \cite{wang2020direct}, UniMP \cite{shi2020masked}, GAT+BoT \cite{wang2021bag} and RevGNN \cite{li2021training}. 

\paragraph{Experimental setup}
For ogbn-arxiv, we utilize 3 AGDN layers with transition matrix of GAT, hidden dimension 256, 3 attention heads, and residual linear connections. For AGDN with BoT, we utilize pseudo-symmetric normalized transition matrix of GAT from BoT. For AGDN with BoT and GIANT-XRT embedding, we utilize our proposed pseudo-symmetric transition matrix of GAT and 2 AGDN layers. For ogbn-proteins, we utilize 6 AGDN layers with the transition matrix of GAT $\overline{\boldsymbol{\mathcal A}}$, hidden dimension 150, 6 attention heads, and residual linear connections. For ogbn-products, we utilize 4 AGDN layers with the transition matrix of GAT, hidden dimension 120, 4 attention heads, and residual linear connections.

\begin{table}[hbt]
\caption{The first part of experimental results on the ogbn-arxiv dataset. Except for AGDN, other results are from their papers or the OGB leaderboard.}
\begin{center}
\begin{tabular}[l]{c|cc|c}
\toprule[1pt]
\multirow{2}{*}{\textbf{Models}} & \multicolumn{2}{c|}{\textbf{Accuracy (\%)}} & \multirow{2}{*}{\textbf{Params}}\\
 & \textbf{Test} & \textbf{Valid} & \\
\midrule
\textbf{GCN}  & 71.74\scriptsize{±0.29} & 73.00\scriptsize{±0.17} & 0.11M \\ 
\textbf{GraphSAGE} & 71.49\scriptsize{±0.27} & 72.77\scriptsize{±0.16} & 0.22M \\
\textbf{DeeperGCN} & 71.92\scriptsize{±0.16} & 72.62\scriptsize{±0.14} & 0.49M \\ 
\textbf{JKNet} & 72.19\scriptsize{±0.21} & 73.35\scriptsize{±0.07} & 0.09M \\
\textbf{DAGNN} & 72.09\scriptsize{±0.25} & 72.90\scriptsize{±0.11} & 0.04M \\
\textbf{GCNII} & 72.74\scriptsize{±0.16} & – & 2.15M \\ 
\textbf{MAGNA} & 72.76\scriptsize{±0.14} & – & – \\
\textbf{Ours (AGDN)} & \underline{\textbf{73.41\scriptsize{±0.25}}} & \underline{\textbf{74.23\scriptsize{±0.13}}} & 1.45M \\ 
\bottomrule[1pt]

\end{tabular}
\end{center}
\label{tab: experiments on ogbn-arxiv}
\end{table}

\begin{table}[!hbt]
\caption{The second part of experimental results on the ogbn-arxiv dataset. Except for AGDN, other results are from their papers or the OGB leaderboard. \ding{172}=BoT, \ding{173}=self-KD, \ding{174}=GIANT-XRT embedding.}
\begin{center}
\setlength{\tabcolsep}{2.5mm}{
    \begin{tabular}{c|cc|c}
    \toprule[1pt]
    \multirow{2}{*}{\textbf{Models}} & \multicolumn{2}{c|}{\textbf{Accuracy(\%)}} & \multirow{2}{*}{\textbf{Params}}\\
     & \textbf{Test} & \textbf{Valid} & \\
        \midrule
        \textbf{UniMP} & 73.11\scriptsize{±0.20} & 74.50\scriptsize{±0.15} & 0.18M \\
        \textbf{GAT+\ding{172}} & 73.91\scriptsize{±0.12} & 75.16\scriptsize{±0.08} & 1.44M \\
        \textbf{RevGAT+\ding{172}} & 74.02\scriptsize{±0.18} & 75.01\scriptsize{±0.10} & 2.10M \\
        \textbf{Ours (AGDN+\ding{172})} & \underline{74.11\scriptsize{±0.12}} & \underline{75.25\scriptsize{±0.05}} & 1.51M \\ 
        \midrule
        \textbf{GAT+\ding{172}+\ding{173}} & 74.16\scriptsize{±0.08} & 75.14\scriptsize{±0.04} & 1.44M \\
        \textbf{RevGAT+\ding{172}+\ding{173}} & 74.26\scriptsize{±0.17} & 74.97\scriptsize{±0.08} & 2.10M \\
        \textbf{Ours (AGDN+\ding{172}+\ding{173})} & \underline{74.31\scriptsize{±0.12}} & \underline{75.22\scriptsize{±0.09}} & 1.51M \\
        \midrule
        \textbf{RevGAT+\ding{172}+\ding{174}} & 75.90\scriptsize{±0.19} & 77.01\scriptsize{±0.09} & 1.30M\\
        \textbf{Ours (AGDN+\ding{172}+\ding{174})} & \underline{76.18\scriptsize{±0.16}} & \underline{\textbf{77.24\scriptsize{±0.06}}} & 1.31M \\
        \midrule
        \textbf{RevGAT+\ding{172}+\ding{173}+\ding{174}} & 76.15\scriptsize{±0.10} & 77.16\scriptsize{±0.09} & 1.30M \\
        \textbf{Ours (AGDN+\ding{172}+\ding{173}+\ding{174})} & \underline{\textbf{76.37\scriptsize{±0.11}}} & \underline{77.19\scriptsize{±0.08}} & 1.31M \\
    \bottomrule[1pt]
    \end{tabular}
    }
    \label{tab: ogbn-arxiv more}
\end{center}
\end{table}

\begin{table}[hbt]
\caption{Experimental results on the ogbn-proteins dataset. DeeperGCN, UniMP, RevGNN, and AGDN are implemented with \textbf{random partition}. GAT is implemented with neighbor sampling. AGDN+BoT is based on the implementation of GAT+BoT, however, labels are not used as model input since they empirically bring no improvements. Except for AGDN, other results are from their papers or the OGB leaderboard.}
\begin{center}
\begin{tabular}[l]{c|cc|c}
\toprule[1pt]
\multirow{2}{*}{\textbf{Models}} & \multicolumn{2}{c|}{\textbf{ROC-AUC(\%)}} & \multirow{2}{*}{\textbf{Params}} \\ 
 & \textbf{Test} & \textbf{Valid} & \\
\midrule
\textbf{GCN}  & 72.51\scriptsize{±0.35} & 79.21\scriptsize{±0.18} & 0.10M \\ 
\textbf{GraphSAGE} & 77.68\scriptsize{±0.20} & 83.34\scriptsize{±0.13} & 0.19M \\ 
\textbf{DeeperGCN} & 85.80\scriptsize{±0.17} & 91.06\scriptsize{±0.16} & 2.37M \\
\textbf{UniMP} & 86.42\scriptsize{±0.08} & 91.75\scriptsize{±0.06} & 1.91M \\
\textbf{GAT+BoT} & 87.65\scriptsize{±0.08} & 92.80\scriptsize{±0.08} & 2.48M \\ 
\textbf{RevGNN-deep} & 87.74\scriptsize{±0.13} & 93.26\scriptsize{±0.06} & 20.03M \\
\textbf{RevGNN-wide} & 88.24\scriptsize{±0.15} & \textbf{94.50\scriptsize{±0.08}} & 68.47M \\ 
\textbf{Ours (AGDN)} & \underline{\textbf{88.65\scriptsize{±0.13}}} & \underline{94.18\scriptsize{±0.05}} & 8.61M \\ 
\bottomrule[1pt]
\end{tabular}
\end{center}
\label{tab: experiments on ogbn-proteins}
\end{table}

\begin{table}[hbt]
\caption{Experimental results on the ogbn-products dataset. GAT, DeeperGCN, and AGDN are implemented with \textbf{random partition}. GraphSAGE and UniMP are implemented with neighbor sampling. Except for AGDN, all results are from their papers or the OGB leaderboard.}
\begin{center}
\begin{tabular}[l]{c|cc|c}
\toprule[1pt]
\multirow{2}{*}{\textbf{Models}} & \multicolumn{2}{c|}{\textbf{Accuracy(\%)}} & \multirow{2}{*}{\textbf{Params}} \\ 
 & \textbf{Test} & \textbf{Valid} & \\
\midrule
\textbf{GCN}  & 75.64\scriptsize{±0.21} & 92.00\scriptsize{±0.03} & 0.10M \\ 
\textbf{GraphSAGE} & 78.50\scriptsize{±0.14} & 92.24\scriptsize{±0.07} & 0.21M \\
\textbf{GraphSAINT} & 80.27\scriptsize{±0.26} & – & 0.33M \\
\textbf{DeeperGCN} & 80.98\scriptsize{±0.20} & 92.38\scriptsize{±0.09} & 0.25M \\
\textbf{SIGN} & 80.52\scriptsize{±0.16} & 92.99\scriptsize{±0.04} & 3.48M \\
\textbf{UniMP} & 82.56\scriptsize{±0.31} & \textbf{93.08\scriptsize{±0.17}} & 1.48M \\
\textbf{RevGNN-112} & 83.07\scriptsize{±0.30} & 92.90\scriptsize{±0.07} & 2.95M \\
\textbf{Ours (AGDN)} & \underline{\textbf{83.34\scriptsize{±0.27}}} & \underline{92.29\scriptsize{±0.10}} & 1.54M \\ 
\bottomrule[1pt]

\end{tabular}
\end{center}
\label{tab: experiments on ogbn-products}
\end{table}

\subsubsection{Results on the ogbn-arxiv dataset}
We split the comparison on the ogbn-arxiv dataset into two parts since several critical tricks contribute significantly to the final performance on this leaderboard. We implement AGDN based on the implementation of GAT+BoT. Firstly, for the first part of the comparison (Table \ref{tab: experiments on ogbn-arxiv}), we disable the options about BoT and compare baselines and AGDN without using labels as input. AGDN outperforms other baselines. Secondly, we adapt BoT, self-KD, and GIANT-XRT for AGDN in the second part of the comparison (Table \ref{tab: ogbn-arxiv more}). We use baselines using labels as input. By progressively applying these tricks, AGDN consistently outperforms GAT and RevGAT. Finally, AGDN+BoT+self-KD+GIANT-XRT achieves a new SOTA performance of 76.37\% with a significant margin. Without GIANT-XRT embedding, AGDN is implemented with 3 layers with hidden dimension 256, and RevGAT is implemented with 5 layers with hidden dimension 256. With more complexity, RevGAT can achieve similar performance to AGDN. With GIANT-XRT embedding, AGDN and RevGAT are implemented with 2 layers with hidden dimension 256. With similar complexity, the margin between RevGAT and AGDN becomes larger.

\subsubsection{Results on the ogbn-proteins and ogbn-products datasets}
Moreover, we evaluate AGDN on larger ogbn-proteins and ogbn-products datasets with the random graph partition technique. For ogbn-proteins, we utilize HC instead of HA. AGDN can achieve a new SOTA result of $88.65\%$, which even outperforms the much more complex and deeper model RevGNN. We only utilize 6 AGDN layers with hidden dimension 150 with 8.61M parameters. RevGNN-wide includes 448 layers with hidden dimension 224 with 68.47M parameters. Furthermore, the inference of AGDN is also conducted on the same GPU card of training with 16Gb memory. In contrast, the inference of RevGNN is conducted on another GPU card with 48Gb memory. 
For ogbn-products, we evaluate AGDN with the random partition. AGDN significantly outperforms other baselines, including RevGNN. AGDN achieves the SOTA performance among models without using labels as input.  

\subsubsection{Runtime}
We report training and inference runtime on the ogbn-proteins dataset in Table \ref{tab:runtime on ogbn-proteins} with the runtime of RevGNNs reported in its paper. This comparison demonstrates that AGDN can outperform RevGNN and cost much less runtime simultaneously. With extended runtime, RevGNN-Deep and RevGNN-Wide cost 2.86Gb and 7.91Gb for training, while AGDN costs 13.67Gb. However, the inference of AGDN is conducted on the same GPU, while the inference of RevGNNs is on another Nvidia RTX A6000 (48Gb) without reporting inference runtime or memory cost.

\begin{table}[!hbt]
\caption{Runtime comparison on the ogbn-proteins dataset with similar Tesla V100 cards.}
\label{tab:runtime on ogbn-proteins}
\small
\begin{center}
    \begin{tabular}{c|ccc}
    \toprule
       \textbf{Model}  & \textbf{Training} & \textbf{Inference} \\
       \midrule
         RevGNN-Deep & 13.5d/2000epochs & – \\
         RevGNN-Wide & 17.1d/2000epochs & – \\
         AGDN & 0.14d/2000epochs & 12s \\
    \bottomrule
    \end{tabular}

\end{center}

\end{table}

\begin{figure}[htb]
    \centering
    \includegraphics[scale=0.48]{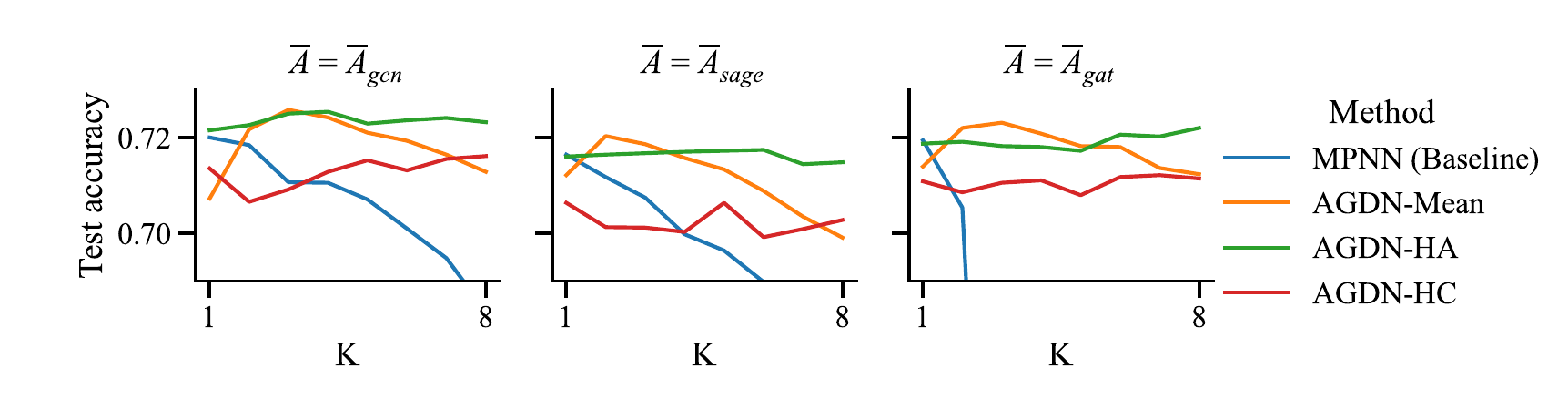}
    \caption{Comparison of AGDN-Mean, AGDN-HA, AGDN-HC with different base models and diffusion hops on the ogbn-arxiv dataset.}
    \label{fig:raw}
\end{figure}

\begin{figure}[htb]
    \centering
    \includegraphics[scale=0.48]{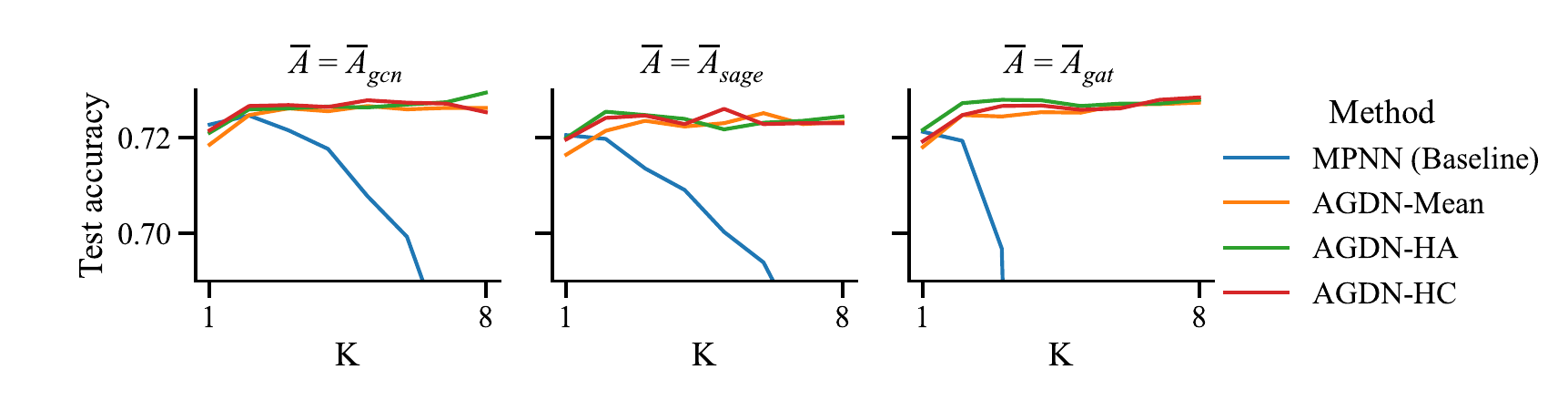}
    \caption{Comparison of AGDN-Mean, AGDN-HA, AGDN-HC with residual linear connections, different base models, and different diffusion hops on the ogbn-arxiv dataset.}
    \label{fig:residual}
\end{figure}

\subsubsection{Over-smoothing and ablation study}
Based on the naive implementation of the official OGB repository, we compare variants of AGDNs with different base models (GCN, GraphSAGE, GAT), under different diffusion depths $K$ ($K=1,2,...,8$), on the ogbn-arxiv dataset. We also evaluate MPNN baselines with related equivalent receptive fields. For example, for $K=4$ in each subgraph of Figure \ref{fig:raw} and Figure \ref{fig:residual}, the associated model in MPNN baseline curve has $3\times4$ layers.
As shown in Figure \ref{fig:raw}, three MPNN baseline curves, especially for GAT, show a distinct over-smoothing problem. AGDN-mean shows quickly rising but then significantly dropping curves, which is also affected by over-smoothing. AGDN-HA and AGDN-HC show much more stable curves. AGDN-HA has similar optimal results to AGDN-mean. However, AGDN-HC cannot even outperform shallow baseline models.
Moreover, we repeat these experiments by adding residual linear connections. As shown in Figure \ref{fig:residual}, all models, including MPNNs (with few layers) and AGDNs, can be improved with residual linear connections. MPNNs still have a distinct over-smoothing problem. However, the over-smoothing problem of AGDN-mean is significantly alleviated. Moreover, AGDN-HC is effectively improved. Three variants of AGDN show similar performance. However, AGDN-HA outperforms AGDN-mean, especially at low $K$. This characteristic is vital in applying complex models on large graphs because we tend to utilize lower $K$ due to the memory limit. This paper selects low $K$ (2 or 3) in other experiments.


\begin{table}[hbt]
    \caption{Ablation study on the ogbn-arxiv, ogbn-proteins and ogbn-products datasets. Due to the space limit, we omit the variances of these scores.}
    \label{tab: ablation study on ogbn}
    \setlength{\tabcolsep}{1.8mm}{
    \begin{center}
    \begin{tabular}{c|cc|cc|cc}
    \toprule[1pt]
    \multirow{3}{*}{\textbf{Models}} & \multicolumn{2}{c|}{\textbf{ogbn-arxiv}} & \multicolumn{2}{c|}{\textbf{ogbn-proteins}} & \multicolumn{2}{c}{\textbf{ogbn-products}} \\
   \cline{2-7} 
         & \multicolumn{2}{c|}{\textbf{Accuracy(\%)}} & \multicolumn{2}{c|}{\textbf{ROC-AUC(\%)}} & \multicolumn{2}{c}{\textbf{Accuracy(\%)}} \\
         & \textbf{Test} & \textbf{Valid} & \textbf{Test} & \textbf{Valid} & \textbf{Test} & \textbf{Valid} \\
    \midrule
    \textbf{GAT} & 72.98 & 74.05 & 88.15 & 93.85 & 81.77 & 91.75 \\
    \textbf{Ours (AGDN)} & \underline{\textbf{73.41}} & \underline{\textbf{74.23}} & \underline{\textbf{88.65}} & \underline{\textbf{94.18}} & \underline{\textbf{83.34}} & \underline{\textbf{92.29}} \\
    \bottomrule[1pt]
    \end{tabular}
    \end{center}
    }

\end{table}

\subsubsection{More ablation study}
We report ablation study between GAT and AGDN on the ogbn-arxiv, ogbn-proteins and ogbn-products datasets in Table \ref{tab: ablation study on ogbn}. We keep all settings the same for GAT and AGDN. We can confirm the significant improvements of AGDN.

\subsection{Task 2: Link Prediction}
\paragraph{Baselines}
For link prediction tasks, we utilize DeepWalk \cite{perozzi2014deepwalk}, Matrix Factorization \cite{menon2011link}, Common Neighbor \cite{liben2007link}, Adamic Adar \cite{adamic2003friends}, Resource Allocation \cite{zhou2009predicting}, GCN \cite{kipf2016semi}, GraphSAGE \cite{hamilton2017inductive}, SEAL \cite{zhang2021labeling} and PLNLP \cite{wang2021pairwise} as baselines. Due to memory limitation, we adapt graph-based sampling techniques including random partition \cite{li2020deepergcn, shi2020masked} for ogbn-proteins and ogbn-products, and GraphSAINT \cite{zeng2019graphsaint} for ogbl-citation2. Some baselines are not implemented for some datasets. Thus we do not report the associated results.

\paragraph{Experimental Setup}
For ogbl-ppa, we utilize 2 AGDN layers with the transition matrix of GAT, hidden dimension 128, 1 attention head, and residual linear connections. For ogbl-ddi, we utilize 2 AGDN layers with the transition matrix of GAT, hidden dimension 512, 1 attention head, and residual linear connections. For ogbl-citation2, we utilize 3 AGDN layers with the transition matrix of GAT, hidden dimension 256, 1 attention head, and residual linear connections. In official OGB baselines, naive cross-entropy loss is used regarding link prediction as binary classification. PLNLP proposes to utilize pairwise AUC loss. We adapt AUC loss on the ogbl-ddi dataset. Note that we only utilize this loss on the ogbl-ddi dataset since it does not improve AGDN on other datasets. We adopt the GraphSAINT technique for AGDN on the ogbl-citation2 dataset. We utilize learnable node embeddings instead of possible node features on the ogbl-ddi (dimension 512) and ogbl-ppa (dimension 128) datasets. Note that we only manually tune a few hyperparameters based on default settings.

\paragraph{Training and evaluation}
We follow the standard training procedure in official OGB baselines, which use an encoder-decoder framework. To emphasize the effect of AGDN, we do not introduce other modifications except pairwise AUC loss. We use the standard data splits and metrics from the official OGB paper for evaluation.

\begin{table}[!hbt]
\caption{Experimental results on the ogbl-ppa dataset.}
\begin{center}
\setlength{\tabcolsep}{1.8mm}{
\begin{tabular}{c|cc|c}
\toprule[1pt]
\multirow{2}*{\textbf{Models}} & \multicolumn{2}{c|}{\textbf{Hits@100(\%)}} & \multirow{2}{*}{\textbf{Params}} \\
 & \textbf{Test} & \textbf{Valid} &  \\
\midrule
\textbf{DeepWalk} & 28.88\scriptsize{±1.53} & - & 150.14M \\
\textbf{Matrix Factorization} & 32.29\scriptsize{±0.94} & 32.28\scriptsize{±4.28} & 147.66M \\
\textbf{Common Neighbor} & 27.65\scriptsize{±0.00} & 28.23\scriptsize{±0.00} & 0 \\
\textbf{Adamic Adar} & 32.45\scriptsize{±0.00} & 32.68\scriptsize{±0.00} & 0 \\
\textbf{Resource Allocation} & \textbf{49.33\scriptsize{±0.00}} & 47.22\scriptsize{±0.00} & 0 \\
\textbf{GCN}  & 18.67\scriptsize{±1.32} & 18.45\scriptsize{±1.40} & 0.28M \\ 
\textbf{GraphSAGE} & 16.55\scriptsize{±2.40} & 17.24\scriptsize{±2.64} & 0.42M\\
\textbf{SEAL} & 48.80\scriptsize{±3.16} & \textbf{51.25\scriptsize{±2.52}} & 0.71M \\
\textbf{PLNLP} & 32.38\scriptsize{±2.58} & - & – \\
\textbf{Ours (AGDN)} & \underline{41.23\scriptsize{±1.59}} & \underline{43.32\scriptsize{±0.92}} & 36.90M \\
\bottomrule[1pt]

\end{tabular}
}
\end{center}
\label{tab: experiments on ogbl-ppa}
\end{table}

\begin{table}[!hbt]
\caption{Experimental results on the ogbl-ddi dataset.}
\begin{center}
\setlength{\tabcolsep}{1.8mm}{
\begin{tabular}{c|cc|c}
\toprule[1pt]
\multirow{2}*{\textbf{Models}} & \multicolumn{2}{c|}{\textbf{Hits@20(\%)}} & \multirow{2}{*}{\textbf{Params}} \\
 & \textbf{Test} & \textbf{Valid} & \\
\midrule
\textbf{DeepWalk} & 22.46\scriptsize{±2.90} & – &  11.54M	 \\
\textbf{Matrix Factorization} & 13.68\scriptsize{±4.75} & 33.70\scriptsize{±2.64} & 1.22M \\
\textbf{Common Neighbor} & 17.73\scriptsize{±0.00} & 9.47\scriptsize{±0.00} & 0 \\
\textbf{Adamic Adar} & 18.61\scriptsize{±0.00} & 9.66\scriptsize{±0.00} & 0 \\
\textbf{Resource Allocation} & 6.23\scriptsize{±0.00} & 7.25\scriptsize{±0.00} & 0 \\
\textbf{GCN} & 37.07\scriptsize{±5.07} & 55.50\scriptsize{±2.08} & 1.29M \\
\textbf{GraphSAGE} & 53.90\scriptsize{±4.74} & 62.62\scriptsize{±0.37} & 1.42M \\
\textbf{SEAL} & 30.56\scriptsize{±3.86} & 28.49\scriptsize{±2.69} & 0.53M \\
\textbf{PLNLP} & 90.88\scriptsize{±3.13} & 82.42\scriptsize{±2.53} & 3.50M \\
\textbf{Ours (AGDN)} & \underline{\textbf{95.38\scriptsize{±0.94}}} & \underline{\textbf{89.43\scriptsize{±2.81}}} & 3.51M \\
\bottomrule[1pt]

\end{tabular}
}
\end{center}
\label{tab: experiments on ogbl-ddi}
\end{table}

\begin{table}[!hbt]
\caption{Experimental results on the ogbl-citation2 dataset.}
\begin{center}
\setlength{\tabcolsep}{1.8mm}{
\begin{tabular}{c|cc|c}
\toprule[1pt]
\multirow{2}{*}{\textbf{Models}} & \multicolumn{2}{c|}{\textbf{MRR(\%)}} & \multirow{2}{*}{\textbf{Params}}\\
 & \textbf{Test} & \textbf{Valid} & \\
\midrule
\textbf{Matrix Factorization} & 51.86\scriptsize{±4.43} & 51.81\scriptsize{±4.36} & 281.11M \\
\textbf{Common Neighbor} & 51.47\scriptsize{±0.00} & 51.19\scriptsize{±0.00} & 0 \\
\textbf{Adamic Adar} & 51.89\scriptsize{±0.00} & 51.67\scriptsize{±0.00} & 0 \\
\textbf{Resource Allocation} & 51.98\scriptsize{±0.00} & 51.77\scriptsize{±0.00} & 0 \\
\textbf{GCN}  & 84.74\scriptsize{±0.31} & 84.79\scriptsize{±0.23} & 0.30M \\ 
\textbf{GraphSAGE} & 82.60\scriptsize{±0.36} & 82.63\scriptsize{±0.33} & 0.46M \\
\textbf{SEAL} & \textbf{87.67\scriptsize{±0.32}} & \textbf{87.57\scriptsize{±0.31}} & 0.26M \\
\textbf{PLNLP} & 84.92\scriptsize{±0.29} & 84.90\scriptsize{±0.31} & 146.51M \\
\textbf{Ours (AGDN)} & \underline{85.49\scriptsize{±0.29}} & \underline{85.56\scriptsize{±0.33}} & 0.31M \\
\bottomrule[1pt]

\end{tabular}
}
\end{center}
\label{tab: experiments on ogbl-citation2}
\end{table}

\begin{table}[!hbt]
\caption{Ablation study on the ogbl-ppa, ogbl-ddi, ogbl-citation2 datasets. Due to the space limit, we omit the variances of these scores.}
\label{tab: ablation study on ogbl}
    \begin{center}
        \begin{tabular}{c|cccccc}
        \toprule[1pt]
        \multirow{3}{*}{\textbf{Models}} & \multicolumn{2}{c}{\textbf{ogbl-ppa}} & \multicolumn{2}{c}{\textbf{ogbl-ddi}} & \multicolumn{2}{c}{\textbf{ogbl-citation2}} \\

         \cline{2-7}& \multicolumn{2}{c}{\textbf{Hits@100(\%)}} & \multicolumn{2}{c}{\textbf{Hits@10(\%)}} & \multicolumn{2}{c}{\textbf{MRR(\%)}} \\
         & \textbf{Test} & \textbf{Valid} & \textbf{Test} & \textbf{Valid} & \textbf{Test} & \textbf{Valid} \\
         \midrule
         \textbf{GAT} & 37.28 & 40.64 & 85.84 & 77.08 & 83.07 & 83.12 \\
         \textbf{Ours (AGDN)} & \underline{\textbf{41.23}} & \underline{\textbf{43.32}} & \underline{\textbf{95.38}} & \underline{\textbf{89.43}} & \underline{\textbf{85.49}} & \underline{\textbf{85.56}} \\
         \bottomrule[1pt]
        \end{tabular}
    \end{center}
\end{table}

\subsubsection{Results}
As shown in Table \ref{tab: experiments on ogbl-ppa}, heuristic methods show significant advantages over GNN methods on the ogbl-ppa dataset. As a GNN architecture modified for link prediction, based on a complicated subgraph extracting and labeling tricks, SEAL achieves similar performance to the best heuristic method. AGDN, based on naive official OGB baseline scripts, outperforms GCN, GraphSAGE, and several heuristic methods. AGDN utilizes learnable node embeddings as model input, bringing additional parameters.

As shown in Table \ref{tab: experiments on ogbl-ddi}, on the ogbl-ddi dataset, GNN methods act better than heuristic methods. With AUC loss, AGDN can achieve 95.38\% Hits@20, the new SOTA result on the ogbl-ddi leaderboard. This dataset is very dense and will make structural patterns meaningless. Thus encoder-decoder GNNs with learnable node embeddings can act much better than SEAL.

As shown in Table \ref{tab: experiments on ogbl-citation2}, on the ogbl-citation2 dataset, GNN methods also act better than heuristic methods. GCN, GraphSAGE, and PLNLP are full-batch trained. Due to our GPU memory limitation (16Gb), we train AGDN with a neighbor sampling technique. We can observe significant performance degradation by comparing full-batch GCN (84.74\% test MRR) and GCN using GraphSAINT (79.85\% test MRR) in the official OGB repository. However, even with GraphSAINT, AGDN still achieves top 3 performance on the whole ogbl-citation2 leaderboard and outperforms full-batch GCN, GraphSAGE, and PLNLP. 

On the ogbl-ddi dataset, AGDN outperforms SEAL and other encoder-decoder GNNs with a significant margin. On the ogbl-ppa and ogbl-citation2 datasets, the margin between AGDN and SEAL is not enormous ($<8$\%) and smaller than other encoder-decoder GNNs. We believe that, with more suitable techniques designed for link prediction, AGDN will contribute more to this task.

\subsubsection{Runtime}
We compare training and inference runtime of AGDN and SEAL on the ogbl-ppa, ogbl-ddi, and ogbl-citation2 datasets in Table \ref{tab:runtime on ogbl}. With similar Tesla V100 GPU cards, AGDN takes significantly less training and inference runtime than SEAL on the ogbl-ppa and ogbl-citation2 datasets. The model architecture of AGDN is more complicated than SEAL. However, the simple encoder-decoder framework is less expressive but much more efficient than SEAL's time-consuming subgraph sampling and labeling trick. On the small ogbl-ddi dataset, where additional techniques in SEAL work much more efficient, AGDN takes more training runtime but still much less inference runtime than SEAL.

\begin{table}[!hbt]
\caption{Runtime comparison on the ogbl-ppa, ogbl-ddi, and ogbl-citation2 datasets with similar Tesla V100 cards.}
\label{tab:runtime on ogbl}
\small
\begin{center}
    \begin{tabular}{cc|ccc}
    \toprule
     \textbf{Dataset} & \textbf{Model}  & \textbf{Training} & \textbf{Inference}  \\
       \midrule
      ogbl-ppa & SEAL & 20h/20epochs & 4h \\
      ogbl-ppa & AGDN & 2.3h/40epochs & 0.06h \\
      ogbl-ddi & SEAL & 0.07h/10epochs & 0.1h \\
      ogbl-ddi & AGDN & 0.8h/2000epochs & 0.3s \\
      ogbl-citation2 & SEAL & 7h/10epochs & 28h \\
      ogbl-citation2 & AGDN & 2.5h/2000epochs & 0.06h \\
    \bottomrule
    \end{tabular}

\end{center}
\end{table}

\subsubsection{Ablation study}
We also conduct an ablation study to demonstrate the effectiveness of AGDN on link prediction tasks. We report the results of both GAT and AGDN (with transition matrix of GAT) on three datasets with the same settings. As shown in Table \ref{tab: ablation study on ogbl}, AGDN significantly improves the link prediction performance of GAT. 

\section{Conclusion}
This paper proposes a feasible and effective evolution path for GNNs. Firstly, we refine and propose Graph Diffusion Networks (GDNs) by replacing the graph convolution operator with an efficient graph diffusion in each GNN layer. Then, we generalize graph diffusion to propose Adaptive Graph Diffusion Networks (AGDNs). We propose two adaptive and scalable mechanisms of computing hop weighting coefficients/matrices. In the spectral domain, AGDNs are more adaptive and flexible than previous graph diffusion-based methods. We evaluate AGDNs and other popular GNNs on node classification and link prediction tasks. The experimental results show that AGDNs can significantly outperform many popular GNNs and even SOTA GNNs (RevGNNs and SEAL). At the same time, AGDNs have considerable overall advantages of complexity and efficiency over SOTA GNNs. Instead of copying huge models from other domains or using simplified architecture, we enlarge the receptive field with moderate complexity and essential architecture. It is valuable for limited computation hardware and time-critical tasks. 

\paragraph{Limits}
As a common issue, it is hard to apply the node-wise or layer-wise neighbor sampling techniques to very deep GNNs, including AGDNs. We must employ additional memory-saving techniques from other models if we want to train a very deep/wide AGDN model with a considerable diffusion depth. Fortunately, AGDNs are compatible with most techniques applied to MPNNs. The effect of position embedding in AGDNs has not been precisely studied. We leave potential memory-saving techniques for AGDNs in future research.
\bibliographystyle{unsrt}  
\bibliography{main}  

\begin{thebibliography}{10}

\bibitem{gilmer2017neural}
Justin Gilmer, Samuel~S Schoenholz, Patrick~F Riley, Oriol Vinyals, and
  George~E Dahl.
\newblock Neural message passing for quantum chemistry.
\newblock {\em arXiv preprint arXiv:1704.01212}, 2017.

\bibitem{kipf2016semi}
Thomas~N Kipf and Max Welling.
\newblock Semi-supervised classification with graph convolutional networks.
\newblock {\em arXiv preprint arXiv:1609.02907}, 2016.

\bibitem{hamilton2017inductive}
Will Hamilton, Zhitao Ying, and Jure Leskovec.
\newblock Inductive representation learning on large graphs.
\newblock In {\em Advances in neural information processing systems}, pages
  1024--1034, 2017.

\bibitem{kipf2016variational}
Thomas~N Kipf and Max Welling.
\newblock Variational graph auto-encoders.
\newblock {\em arXiv preprint arXiv:1611.07308}, 2016.

\bibitem{chen2018fastgcn}
Jie Chen, Tengfei Ma, and Cao Xiao.
\newblock Fastgcn: fast learning with graph convolutional networks via
  importance sampling.
\newblock {\em arXiv preprint arXiv:1801.10247}, 2018.

\bibitem{fout2017protein}
Alex Fout, Jonathon Byrd, Basir Shariat, and Asa Ben-Hur.
\newblock Protein interface prediction using graph convolutional networks.
\newblock In {\em Advances in neural information processing systems}, pages
  6530--6539, 2017.

\bibitem{yu2021deep}
Le~Yu, Bowen Du, Xiao Hu, Leilei Sun, Liangzhe Han, and Weifeng Lv.
\newblock Deep spatio-temporal graph convolutional network for traffic accident
  prediction.
\newblock {\em Neurocomputing}, 423:135--147, 2021.

\bibitem{li2021traffic}
Wei Li, Xin Wang, Yiwen Zhang, and Qilin Wu.
\newblock Traffic flow prediction over muti-sensor data correlation with graph
  convolution network.
\newblock {\em Neurocomputing}, 427:50--63, 2021.

\bibitem{yin2021multi}
Xueyan Yin, Genze Wu, Jinze Wei, Yanming Shen, Heng Qi, and Baocai Yin.
\newblock Multi-stage attention spatial-temporal graph networks for traffic
  prediction.
\newblock {\em Neurocomputing}, 428:42--53, 2021.

\bibitem{li2018deeper}
Qimai Li, Zhichao Han, and Xiao-Ming Wu.
\newblock Deeper insights into graph convolutional networks for semi-supervised
  learning.
\newblock {\em arXiv preprint arXiv:1801.07606}, 2018.

\bibitem{wang2019improving}
Guangtao Wang, Rex Ying, Jing Huang, and Jure Leskovec.
\newblock Improving graph attention networks with large margin-based
  constraints.
\newblock {\em arXiv preprint arXiv:1910.11945}, 2019.

\bibitem{liu2020towards}
Meng Liu, Hongyang Gao, and Shuiwang Ji.
\newblock Towards deeper graph neural networks.
\newblock In {\em Proceedings of the 26th ACM SIGKDD International Conference
  on Knowledge Discovery \& Data Mining}, pages 338--348, 2020.

\bibitem{li2019deepgcns}
Guohao Li, Matthias Muller, Ali Thabet, and Bernard Ghanem.
\newblock Deepgcns: Can gcns go as deep as cnns?
\newblock In {\em Proceedings of the IEEE International Conference on Computer
  Vision}, pages 9267--9276, 2019.

\bibitem{li2020deepergcn}
Guohao Li, Chenxin Xiong, Ali Thabet, and Bernard Ghanem.
\newblock Deepergcn: All you need to train deeper gcns.
\newblock {\em arXiv preprint arXiv:2006.07739}, 2020.

\bibitem{chen2020simple}
Ming Chen, Zhewei Wei, Zengfeng Huang, Bolin Ding, and Yaliang Li.
\newblock Simple and deep graph convolutional networks.
\newblock In {\em International Conference on Machine Learning}, pages
  1725--1735. PMLR, 2020.

\bibitem{li2021training}
Guohao Li, Matthias M{\"u}ller, Bernard Ghanem, and Vladlen Koltun.
\newblock Training graph neural networks with 1000 layers.
\newblock In {\em International conference on machine learning}, pages
  6437--6449. PMLR, 2021.

\bibitem{klicpera2018predict}
Johannes Klicpera, Aleksandar Bojchevski, and Stephan G{\"u}nnemann.
\newblock Predict then propagate: Graph neural networks meet personalized
  pagerank.
\newblock {\em arXiv preprint arXiv:1810.05997}, 2018.

\bibitem{rossi2020sign}
Emanuele Rossi, Fabrizio Frasca, Ben Chamberlain, Davide Eynard, Michael
  Bronstein, and Federico Monti.
\newblock Sign: Scalable inception graph neural networks.
\newblock {\em arXiv preprint arXiv:2004.11198}, 2020.

\bibitem{klicpera2019diffusion}
Johannes Klicpera, Stefan Wei{\ss}enberger, and Stephan G{\"u}nnemann.
\newblock Diffusion improves graph learning.
\newblock In {\em Advances in Neural Information Processing Systems}, pages
  13354--13366, 2019.

\bibitem{zhao2019pairnorm}
Lingxiao Zhao and Leman Akoglu.
\newblock Pairnorm: Tackling oversmoothing in gnns.
\newblock {\em arXiv preprint arXiv:1909.12223}, 2019.

\bibitem{rong2019dropedge}
Yu~Rong, Wenbing Huang, Tingyang Xu, and Junzhou Huang.
\newblock Dropedge: Towards deep graph convolutional networks on node
  classification.
\newblock {\em arXiv preprint arXiv:1907.10903}, 2019.

\bibitem{feng2020graph}
Wenzheng Feng, Jie Zhang, Yuxiao Dong, Yu~Han, Huanbo Luan, Qian Xu, Qiang
  Yang, Evgeny Kharlamov, and Jie Tang.
\newblock Graph random neural networks for semi-supervised learning on graphs.
\newblock {\em Advances in neural information processing systems},
  33:22092--22103, 2020.

\bibitem{hu2020open}
Weihua Hu, Matthias Fey, Marinka Zitnik, Yuxiao Dong, Hongyu Ren, Bowen Liu,
  Michele Catasta, and Jure Leskovec.
\newblock Open graph benchmark: Datasets for machine learning on graphs.
\newblock {\em arXiv preprint arXiv:2005.00687}, 2020.

\bibitem{xu2018representation}
Keyulu Xu, Chengtao Li, Yonglong Tian, Tomohiro Sonobe, Ken-ichi Kawarabayashi,
  and Stefanie Jegelka.
\newblock Representation learning on graphs with jumping knowledge networks.
\newblock {\em arXiv preprint arXiv:1806.03536}, 2018.

\bibitem{page1999pagerank}
Lawrence Page, Sergey Brin, Rajeev Motwani, and Terry Winograd.
\newblock The pagerank citation ranking: Bringing order to the web.
\newblock Technical report, Stanford InfoLab, 1999.

\bibitem{kondor2002diffusion}
Risi~Imre Kondor and John Lafferty.
\newblock Diffusion kernels on graphs and other discrete structures.
\newblock In {\em Proceedings of the 19th international conference on machine
  learning}, volume 2002, pages 315--22, 2002.

\bibitem{xu2020graph}
Bingbing Xu, Huawei Shen, Qi~Cao, Keting Cen, and Xueqi Cheng.
\newblock Graph convolutional networks using heat kernel for semi-supervised
  learning.
\newblock {\em arXiv preprint arXiv:2007.16002}, 2020.

\bibitem{berberidis2018adaptive}
Dimitris Berberidis, Athanasios~N Nikolakopoulos, and Georgios~B Giannakis.
\newblock Adaptive diffusions for scalable learning over graphs.
\newblock {\em IEEE Transactions on Signal Processing}, 67(5):1307--1321, 2018.

\bibitem{chen2013adaptive}
Siheng Chen, Aliaksei Sandryhaila, Jos{\'e}~MF Moura, and Jelena
  Kova{\v{c}}evi{\'c}.
\newblock Adaptive graph filtering: Multiresolution classification on graphs.
\newblock In {\em 2013 IEEE Global Conference on Signal and Information
  Processing}, pages 427--430. IEEE, 2013.

\bibitem{abu2018watch}
Sami Abu-El-Haija, Bryan Perozzi, Rami Al-Rfou, and Alexander~A Alemi.
\newblock Watch your step: Learning node embeddings via graph attention.
\newblock In {\em Advances in Neural Information Processing Systems}, pages
  9180--9190, 2018.

\bibitem{wu2019simplifying}
Felix Wu, Tianyi Zhang, Amauri Holanda~de Souza~Jr, Christopher Fifty, Tao Yu,
  and Kilian~Q Weinberger.
\newblock Simplifying graph convolutional networks.
\newblock {\em arXiv preprint arXiv:1902.07153}, 2019.

\bibitem{zhu2021simple}
Hao Zhu and Piotr Koniusz.
\newblock Simple spectral graph convolution.
\newblock In {\em International Conference on Learning Representations}, 2021.

\bibitem{atwood2016diffusion}
James Atwood and Don Towsley.
\newblock Diffusion-convolutional neural networks.
\newblock In {\em Advances in neural information processing systems}, pages
  1993--2001, 2016.

\bibitem{du2017topology}
Jian Du, Shanghang Zhang, Guanhang Wu, Jos{\'e}~MF Moura, and Soummya Kar.
\newblock Topology adaptive graph convolutional networks.
\newblock {\em arXiv preprint arXiv:1710.10370}, 2017.

\bibitem{wang2020direct}
Guangtao Wang, Rex Ying, Jing Huang, and Jure Leskovec.
\newblock Direct multi-hop attention based graph neural network.
\newblock {\em arXiv preprint arXiv:2009.14332}, 2020.

\bibitem{abu2019mixhop}
Sami Abu-El-Haija, Bryan Perozzi, Amol Kapoor, Nazanin Alipourfard, Kristina
  Lerman, Hrayr Harutyunyan, Greg~Ver Steeg, and Aram Galstyan.
\newblock Mixhop: Higher-order graph convolutional architectures via sparsified
  neighborhood mixing.
\newblock {\em arXiv preprint arXiv:1905.00067}, 2019.

\bibitem{abu2020n}
Sami Abu-El-Haija, Amol Kapoor, Bryan Perozzi, and Joonseok Lee.
\newblock N-gcn: Multi-scale graph convolution for semi-supervised node
  classification.
\newblock In {\em uncertainty in artificial intelligence}, pages 841--851.
  PMLR, 2020.

\bibitem{wang2021bag}
Yangkun Wang.
\newblock Bag of tricks of semi-supervised classification with graph neural
  networks.
\newblock {\em arXiv preprint arXiv:2103.13355}, 2021.

\bibitem{zhang2019your}
Linfeng Zhang, Jiebo Song, Anni Gao, Jingwei Chen, Chenglong Bao, and Kaisheng
  Ma.
\newblock Be your own teacher: Improve the performance of convolutional neural
  networks via self distillation.
\newblock In {\em Proceedings of the IEEE/CVF International Conference on
  Computer Vision}, pages 3713--3722, 2019.

\bibitem{chien2021node}
Eli Chien, Wei-Cheng Chang, Cho-Jui Hsieh, Hsiang-Fu Yu, Jiong Zhang, Olgica
  Milenkovic, and Inderjit~S Dhillon.
\newblock Node feature extraction by self-supervised multi-scale neighborhood
  prediction.
\newblock {\em arXiv preprint arXiv:2111.00064}, 2021.

\bibitem{zhang2021labeling}
Muhan Zhang, Pan Li, Yinglong Xia, Kai Wang, and Long Jin.
\newblock Labeling trick: A theory of using graph neural networks for
  multi-node representation learning.
\newblock {\em Advances in Neural Information Processing Systems},
  34:9061--9073, 2021.

\bibitem{zhang2018link}
Muhan Zhang and Yixin Chen.
\newblock Link prediction based on graph neural networks.
\newblock {\em Advances in neural information processing systems}, 31, 2018.

\bibitem{ng2001spectral}
Andrew Ng, Michael Jordan, and Yair Weiss.
\newblock On spectral clustering: Analysis and an algorithm.
\newblock {\em Advances in neural information processing systems}, 14, 2001.

\bibitem{velivckovic2017graph}
Petar Veli{\v{c}}kovi{\'c}, Guillem Cucurull, Arantxa Casanova, Adriana Romero,
  Pietro Lio, and Yoshua Bengio.
\newblock Graph attention networks.
\newblock {\em arXiv preprint arXiv:1710.10903}, 2017.

\bibitem{shi2020masked}
Yunsheng Shi, Zhengjie Huang, Shikun Feng, and Yu~Sun.
\newblock Masked label prediction: Unified massage passing model for
  semi-supervised classification.
\newblock {\em arXiv preprint arXiv:2009.03509}, 2020.

\bibitem{perozzi2014deepwalk}
Bryan Perozzi, Rami Al-Rfou, and Steven Skiena.
\newblock Deepwalk: Online learning of social representations.
\newblock In {\em Proceedings of the 20th ACM SIGKDD international conference
  on Knowledge discovery and data mining}, pages 701--710, 2014.

\bibitem{menon2011link}
Aditya~Krishna Menon and Charles Elkan.
\newblock Link prediction via matrix factorization.
\newblock In {\em Joint european conference on machine learning and knowledge
  discovery in databases}, pages 437--452. Springer, 2011.

\bibitem{liben2007link}
David Liben-Nowell and Jon Kleinberg.
\newblock The link-prediction problem for social networks.
\newblock {\em Journal of the American society for information science and
  technology}, 58(7):1019--1031, 2007.

\bibitem{adamic2003friends}
Lada~A Adamic and Eytan Adar.
\newblock Friends and neighbors on the web.
\newblock {\em Social networks}, 25(3):211--230, 2003.

\bibitem{zhou2009predicting}
Tao Zhou, Linyuan L{\"u}, and Yi-Cheng Zhang.
\newblock Predicting missing links via local information.
\newblock {\em The European Physical Journal B}, 71(4):623--630, 2009.

\bibitem{wang2021pairwise}
Zhitao Wang, Yong Zhou, Litao Hong, Yuanhang Zou, and Hanjing Su.
\newblock Pairwise learning for neural link prediction.
\newblock {\em arXiv preprint arXiv:2112.02936}, 2021.

\bibitem{zeng2019graphsaint}
Hanqing Zeng, Hongkuan Zhou, Ajitesh Srivastava, Rajgopal Kannan, and Viktor
  Prasanna.
\newblock Graphsaint: Graph sampling based inductive learning method.
\newblock {\em arXiv preprint arXiv:1907.04931}, 2019.

\end{thebibliography}


\end{document}